\documentclass[10pt,twocolumn,letterpaper]{article}
\usepackage[pagenumbers]{cvpr} 
\usepackage[dvipsnames]{xcolor}

\usepackage{times}

\definecolor{cvprblue}{rgb}{0.21,0.49,0.74}
\usepackage[pagebackref,breaklinks,colorlinks,citecolor=cvprblue]{hyperref}

\usepackage[utf8]{inputenc} 
\usepackage[T1]{fontenc}    
\usepackage{url}            
\usepackage{booktabs}       
\usepackage{amsfonts}       
\usepackage{nicefrac}       
\usepackage{microtype}      
\usepackage{xcolor}         
\usepackage{bbding}         
\usepackage{graphicx}
\usepackage{tabularx}       
\usepackage{comment}
\usepackage{color}

\usepackage{amsmath}
\usepackage{amssymb}
\usepackage{float}
\usepackage[symbol]{footmisc}
\usepackage{multirow}

\usepackage[capitalize]{cleveref}
\crefname{section}{Sec.}{Secs.}
\Crefname{section}{Section}{Sections}
\Crefname{table}{Table}{Tables}
\crefname{table}{Tab.}{Tabs.}

\usepackage{hyphenat}
\usepackage{soul,color}
\usepackage{amsopn}

\definecolor{Red}{cmyk}{0,1,1,0}
\definecolor{Green}{cmyk}{1,0,1,0}
\definecolor{Cyan}{cmyk}{1,0,0,0}
\definecolor{Purple}{cmyk}{0.45,0.86,0,0}
\definecolor{Rosolic}{cmyk}{0.00,1.00,0.50,0}
\definecolor{Blue}{cmyk}{1.00,1.00,0.00,0}
\definecolor{BlueViolet}{cmyk}{0.86,0.91,0,0.04}
\definecolor{NavyBlue}{cmyk}{0.94,0.54,0,0}

\newcommand{\hidden}[1]{{\color{NavyBlue}}}

\title{ShapeGPT: 3D Shape Generation with A Unified Multi-modal Language Model}

\author{Fukun Yin$^{1,2}$\thanks{This work was done when Fukun Yin, Biao Jiang, and Zibo Zhao were Research Interns at Tencent.} \quad Xin Chen$^{2}$\thanks{Project Lead} \quad Chi Zhang$^{2}$ \quad Biao Jiang$^{1,2}$ \quad Zibo Zhao$^{2,3}$ \quad Jiayuan Fan$^{1}$ \\Gang Yu$^{2}$ \quad Taihao Li$^{4}$ \quad Tao Chen$^{1}$\thanks{Corresponding author.}
\\ $^{1}$Fudan University \quad\quad\quad$^{2}$Tencent \quad\quad\quad$^{3}$ShanghaiTech University \quad\quad\quad$^{4}$Zhejiang Lab\\
\tt \small \textbf{\href{https://github.com/OpenShapeLab/ShapeGPT}{https://github.com/OpenShapeLab/ShapeGPT}}}

\begin{document}

\makeatletter
\let\@oldmaketitle\@maketitle%
\renewcommand{\@maketitle}{\@oldmaketitle%
 \centering
    \includegraphics[width=1\textwidth]{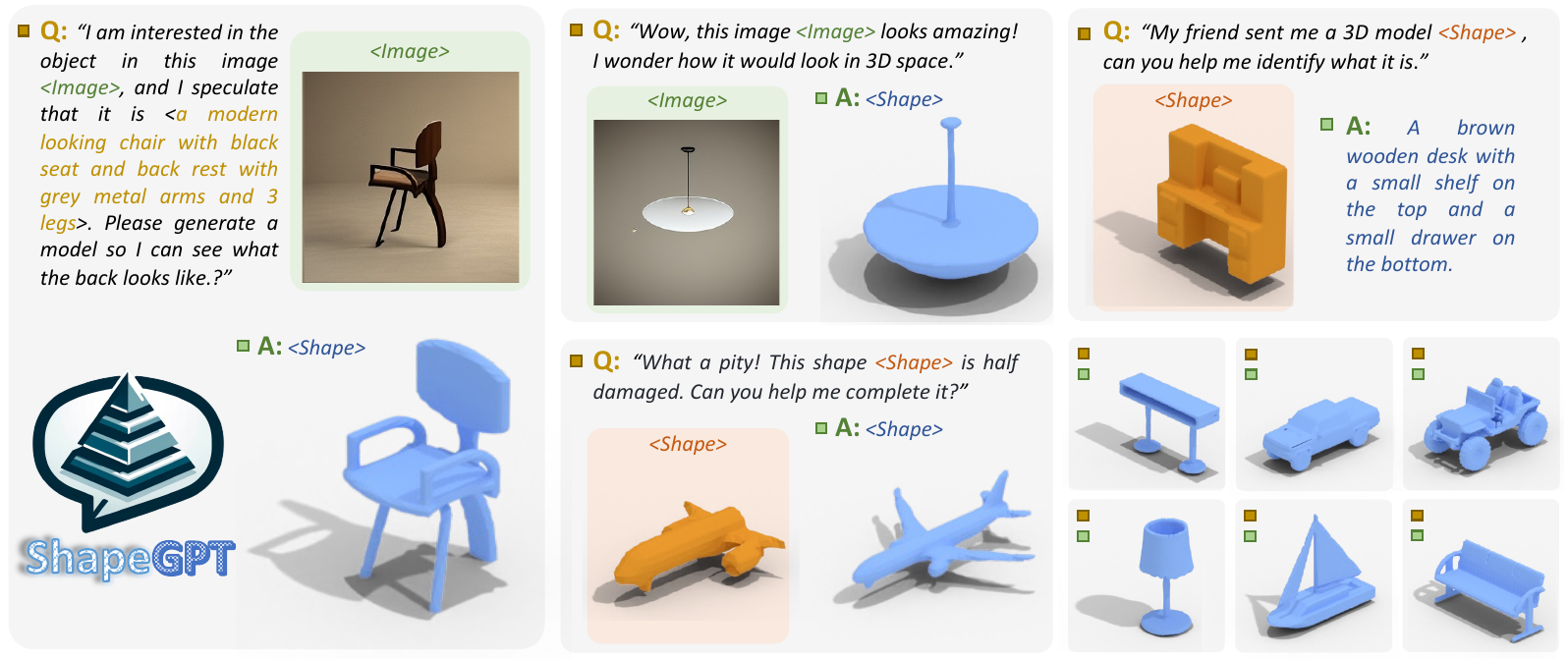}

     \captionof{figure}{\textbf{Illustrative Instances of ShapeGPT.} We present ShapeGPT, a unified generative framework for various shape-centric multimodal tasks, according to the provided instructions without perceptible transition in task handling. The blue shapes are our generated results. }

    \label{fig:teaser}
    \bigskip}                   %
\makeatother

\maketitle

\newcommand{\fix}{\marginpar{FIX}}
\newcommand{\new}{\marginpar{NEW}}

\begin{abstract}
The advent of large language models, enabling flexibility through instruction-driven approaches, has revolutionized many traditional generative tasks, but large models for 3D data, particularly in comprehensively handling 3D shapes with other modalities, are still under-explored.
By achieving instruction-based shape generations, versatile multimodal generative shape models can significantly benefit various fields like 3D virtual construction and network-aided design.
In this work, we present ShapeGPT, a shape-included multi-modal framework to leverage strong pre-trained language models to address multiple shape-relevant tasks.
Specifically, ShapeGPT employs a ``word-sentence-paragraph'' framework to discretize continuous shapes into shape words, further assembles these words for shape sentences, as well as integrates shape with instructional text for multi-modal paragraphs.
To learn this shape-language model, we use a three-stage training scheme, including shape representation, multimodal alignment, and instruction-based generation, to align shape-language codebooks and learn the intricate correlations among these modalities.
Extensive experiments demonstrate that ShapeGPT achieves comparable performance across shape-relevant tasks, including text-to-shape, shape-to-text, shape completion, and shape editing.

\end{abstract}

\section{Introduction}
\label{intro}

\hspace{3mm} The domain of multimodal learning has recently experienced a remarkable transformation. This shift has been primarily attributed to the rise of large-scale pre-trained large language models (LLMs), such as GPT~\cite{radford2018gpt,radford2019gpt2,brown2020gpt3,ouyang2022instructgpt}, BERT~\cite{devlin2018bert}, and T5~\cite{raffel2020t5,chung2022flant5}.
These advancements have expanded the scope of multimodal learning to encompass various modalities, ranging from language~\cite{zhang2022opt,touvron2023llama} and image~\cite{radford2021clip,wang2022beit,li2022blip,xu2021vlm,alayrac2022flamingo} to other innovative modalities~\cite{youwang2022clipactor,mohammad2022clipmesh, jiang2023motiongpt}.
Nevertheless, the exploration of a comprehensive pre-trained model for 3D data, particularly 3D shape and other multimodal data, remains insufficient. This shape-aware multimodal generative model, capable of supporting numerous shape-relevant tasks through instructions, should enrich diverse fields like virtual construction~\cite{luo20233d}, gaming~\cite{kargas2019using}, network-aided design~\cite{Gumeli_2022_CVPR}, and 3D printing~\cite{brion2022generalisable}.

Previous research in the domain of 3D shapes has explored many tasks, including shape generation~\cite{mittal2022autosdf,zhao2023michelangelo,xie2019pix2vox,ibing2023octree}, shape captioning~\cite{han2020shapecaptioner}, and shape completion/editing\cite{wu2020multimodal,kim2023pointinverter}, with a recent trend shifting towards the incorporation of text instructions to enhance flexibility in shape generation. Notable examples of this approach include ShapeCrafter~\cite{fu2022shapecrafter} and SDFusion~\cite{cheng2022sdfusion}, which leverage the capabilities of pre-trained image-text models like CLIP~\cite{radford2021learning} for conditional text-based shape generation.
However, a common limitation across these works is the independent solution of tasks, typically favoring a singular modality translation approach.
Furthermore, the specificity of the training datasets often restricts their versatility, limiting the model's ability to adapt across a variety of tasks or handle diverse input combinations. These limitations stem from a lack of holistic understanding of the interplay between 3D shapes and other modalities and the absence of mutual benefits across shape datasets.

Given these constraints in existing research, our motivation stems from the aspiration to devise a multi-modal model that exhibits versatility across tasks, seamlessly integrating and outputting varied modalities, all under the guidance of language instructions. At the core of our model is  an LLM that comprehends language instructions and orchestrates information from multi-modality data, including shape representations, images, and texts.
However, constructing such a robust multimodal LLM presents challenges, including the development of multimodal neural representations compatible with language models and the establishment of a unified multi-task framework to effectively model relationships between different modalities.

In response to these challenges, we propose ShapeGPT, a unified shape-included multi-modal framework that leverages the strong language understanding and generation abilities of pre-trained LLMs to address multiple shape-related tasks. Inspired by pioneering language-driven vision models like Unified-IO~\cite{lu2022unified} and MotionGPT~\cite{jiang2023motiongpt}, our approach tokenizes information from shape data and integrates them into a language codebook through a shape-specific vector quantized variational autoencoder (VQ-VAE) ~\cite{van2017neural}. This process converts 3D shapes into a series of shape tokens, which are then processed by a pre-trained language model to learn the underlying grammar and syntax of encoded shapes and their relationship with textual descriptions. For image inputs, we employ a visual transformer and a visual-language perceiver to efficiently integrate image information into the language models.
To enable the effective learning of this multi-modal framework, we design a three-stage training scheme.
We first pre-train the shape-language model on 3D datasets to learn basic shape generation and shape captioning in the language model.  For multi-modal learning, we fine-tune the shape-language model on the multi-modal dataset, which contains textual descriptions, rendered images, and shape data, to learn the correlation between the three modalities.
Our extensive experiments demonstrate that ShapeGPT achieves state-of-the-art performance on a variety of tasks.

We summarize our contributions as follows: 
(1) We propose a uniform shape-language generative pre-trained model, ShapeGPT, which involves multi-modal condition inputs, introduces natural language models into shape-relevant generation, and performs diverse shape tasks with a single model.
(2) We introduce a shape-aware multi-modal training scheme with instructions, to learn from task feedback and produce promising results through prompts.
(3) We propose a general shape benchmark for multi-task evaluation, wherein ShapeGPT achieves competitive performance across diverse tasks, including image-to-shape, text-to-shape, shape-to-text, multi-modal-to-shape, shape completion, and shape editing, with all available codes and data.

\section{Related Work}
\label{relatedwork}

\begin{figure*}[t]
\centering
\includegraphics[width=\linewidth]{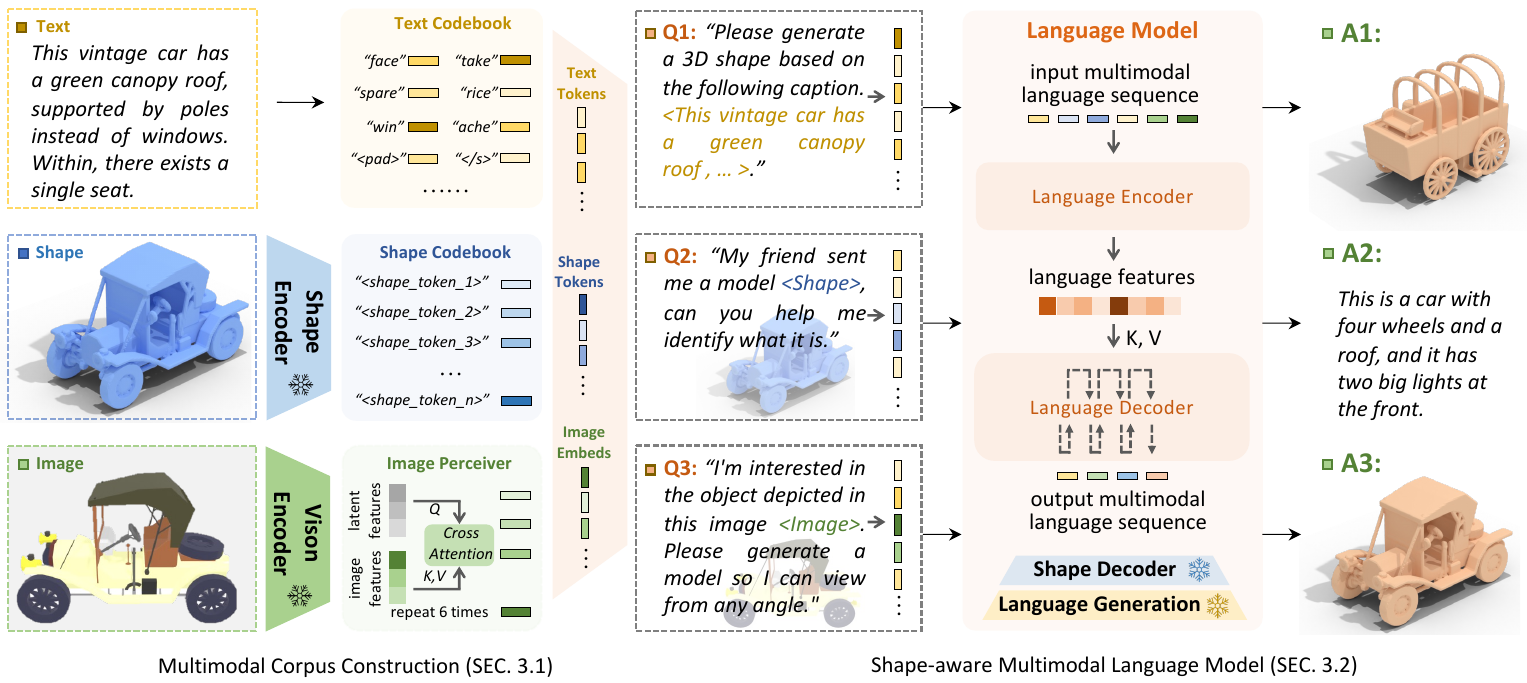}
\vspace{-12pt}
\caption{The overview of the framework. ShapeGPT consists of two parts, \textit{Multimodal Corpus Construction} to tokenize multimodal inputs for corpus collection (\cref{sec:method:corpus}) and \textit{Shape-aware Multimodal Language Model} to comprehend vision-shape-language grammar (\cref{sec:method:lm}) aiming at diverse shape-relevant generations, including image-to-shape, text-to-shape, shape-to-text, shape completion, and shape editing.}
\vspace{-10pt}
\label{fig:pipeline}
\end{figure*}

\hspace{3mm} \textbf{3D Shape Generation and Representation.} Recent works involves generating diverse and high-field 3D shapes using multi-modal input conditions, such as text~\cite{cheng2022sdfusion, zhao2023michelangelo}, images~\cite{xie2019pix2vox,mittal2022autosdf}, categories~\cite{zhang20233dshape2vecset}, point clouds~\cite{zhou20213d} through diverse 3D representation, which includes point clouds~\cite{zhou20213d,melas2023pc2}, voxels~\cite{sella2023vox,wang2021voxel}, meshes~\cite{wen2022pixel2mesh++,sanghi2022clip}, SDF~\cite{cheng2022sdfusion,mittal2022autosdf}, latent vectors~\cite{zhao2023michelangelo, zhang20233dshape2vecset}, and neural fields~\cite{yin2022coordinates,yariv2021volume}.
Text-to-shape~\cite{cheng2022sdfusion, zhao2023michelangelo} and image-to-shape~\cite{xie2019pix2vox,mittal2022autosdf} are two important conditional shape generation tasks.
ShapeCrafter~\cite{fu2022shapecrafter} focuses on text condition and learns a recursive shape generative model leveraging on the text encoder of CLIP~\cite{radford2021clip}, while
SDFusion~\cite{cheng2022sdfusion} design a diffusion-based multi-modal shape generation framework that can fuse text, images, and partial shapes.
3DShape2VecSet~\cite{zhang20233dshape2vecset} involves image conditions also incorporates a new representation of 3D shapes encoded as latent vectors for diffusion models, also leveraging CLIP~\cite{radford2021clip} as conditional text or image encoders.
Although they show promising results in various shape tasks, most above works are limited to a single model to handle multiple tasks.
We thus propose a unified shape-aware multi-modal generative model that leverages the strong language models to achieve diverse shape-relevant generation tasks.

\textbf{Language Models and Multi-Modal.}

Recent advancements in Large Language Models (LLMs)~\cite{dai2019transformer,raffel2020t5,brown2020gpt3,zhang2022opt,touvron2023llama, chung2022flant5, gpt2, vicuna, gpt4} have been pivotal in elevating natural language processing, powered by extensive datasets and significant increases in model size.
These models exhibit advanced comprehension and generation capabilities, particularly ChatGPT and GPT4, which employ a unified framework that transforms all language-based tasks into a text-to-text format.
Subsequent research of multi-modal models~\cite{li2022blip,huang2023language,li2023blip2} has opened new avenues in processing text alongside other modalities, including images~\cite{li2022blip,huang2023language,girdhar2023imagebind}, audio~\cite{guzhov2022audioclip,girdhar2023imagebind}, motion~\cite{jiang2023motiongpt} and videos~\cite{xu2021videoclip}.
These models integrate multiple forms of data, offering enriched contexts and deeper understanding.
For instance, advancements in image-language models~\cite{li2023blip2, liu2023llava, ye2023mplug, zhu2023minigpt4} have significantly improved the synergy between visual and textual information.
However, despite these significant strides, the exploration of multi-modal language models capable of handling 3D shapes is still in its early stages. We thus focus on generative multi-modal models for instructed shape generation.

\textbf{Shape-language Pre-training.} 
Current text-to-shape generation methods~\cite{cheng2022sdfusion, zhao2023michelangelo, zhang20233dshape2vecset} mostly utilize a tag-to-shape approach, processing textual descriptions to generate desired shapes.
While effective in generating shapes from text, these methods cannot typically interpret user-specific instructions, similar to the limitations observed in InstructGPT~\cite{ouyang2022instructgpt}.
Meanwhile, advanced language models like GPTs and llama~\cite{touvron2023llama} have made significant ability to switch between various language-based tasks.
However, their application in shape-related tasks remains largely unexplored.
To bridge this gap, we present ShapeGPT, a unified shape-language generative pre-trained model to achieve shape-relevant generation tasks.

\section{Method}
\label{method}

\hspace{3mm} To realize a multi-modal, instruction-based shape generative model, we propose a unified language-shape framework known as ShapeGPT. As illustrated in \cref{fig:pipeline}, ShapeGPT encompasses the construction of \textbf{a multimodal corpus construction} (\cref{sec:method:corpus}) that parses raw inputs such as text, images, and shapes into 
discrete multimodal corpus collection
with a predefined format and annotation, and \textbf{a shape-aware language model} (\cref{sec:method:lm}) to comprehend grammar and syntax of different modalities centered around shapes, and design a multi-task natural language generation module aimed at equipping instructions with the appropriate format derived from the multi-modal corpus, thereby accomplishing a variety of tasks. Adhering to the word-sentence-communication paradigm of human language acquisition, we adopt a \textbf{a three-stage training strategy} (\cref{sec:method: strategy}) including corpus pre-training and language generation, multimodal shape-aware model training, and multi-task instruction fine-tuning.

We first train shape Variational Autoencoders to discretize the input shape $s$ into token sequences $z_s = \{z_s^i\}_{i=0}^{l_s-1}$ via the shape encoder $\mathcal{S}_e$ and further design shape-related grammar and syntax to form shape words $w_s = \{w_s^i\}_{i=0}^{l_s-1}$ and shape sentences $s_s = \{s_s^i\}_{i=0}^{l_s+1}$, employ a pre-trained image encoder $\mathcal{P}_e$ to encode images $p$ into localized embeddings $e_p = \{e_p^i\}_{i=0}^{l_p-1}$, and segment the input caption $c$ into discrete tokens $z_c = \{z_c^i\}_{i=0}^{l_c-1}$, where $l_s$, $l_p$, $l_c$ represent the lengths of the shape, image, and caption sequences, respectively.
Then, we build a unified multitask natural language framework based on the foundation of the multimodal corpus $z_s, e_p, z_c$ to encode the multimodal data into instruction $\mathcal{I}$ and expected output $\mathcal{O}$, thereby obtaining uniformly formatted shape-related question $\mathcal{Q}=\{q^i\}_{i=0}^{l_q-1}=\mathcal{I}.fill\{w_s, e_p, z_c\}$ and answer $\mathcal{A}=\{a^i\}_{i=0}^{l_a-1}=\mathcal{O}.fill\{w_s, e_p, z_c\}$ pairs. ShapeGPT thus generates shape sequences and textual sequences, enabling the accomplishment of various shape-related tasks, such as multimodal shape generation, shape caption, shape reasoning, shape editing.

\subsection{Multimodal Corpus Construction}
\label{sec:method:corpus}

\hspace{3mm} Given that both shape and image data are high-dimensional and continuous, it is necessary to initially encode them into low-dimensional discrete tokens that can be comprehended by the language model. Consequently, we pre-train a shape quantization variational autoencoder (VQ-VAE) and employ an image feature encoder to construct the multimodal corpus.

In addressing a specific 3D shape, this shape is first transformed into a Signed Distance Function (SDF) representation $s$ that is more sensitive to topological changes. 
Then, discretization is achieved through a 3D Vector Quantised-Variational AutoEncoder (3D VQ-VAE), comprising a shape encoder $\mathcal{S}e$, a shape decoder $\mathcal{S}_d$, and a shape vector quantization module $\mathcal{S}_{vq}$.
Initially, the encoder $\mathcal{S}_e$ converts the continuous SDF representation $s$ into a latent space, which is then unfolded into a one-dimensional sequence akin to natural language, following the x-y-z axis order. 
The vector quantization module $\mathcal{S}_{vq}$ then maps this latent space z with the closest vector in the codebook $cb=\{cb_j\}_{j=0}^{m-1}$, where $m$ denotes the number of embedding vectors, resulting in a discretized sequence of codebook entries $z_s$. 
The codebook is a learnable embedding, which is simultaneously optimized during our 3D VQ-VAE training process, effectively capturing representative features of 3D shapes. The decoder $\mathcal{S}_d$ reshapes the discretized sequence into a high-dimensional representation and maps it back to the continuous SDF representation $\hat{s}$.

\cref{eq:loss:vq} represents the comprehensive loss function of the 3D VQ-VAE, which comprises three components. The first term corresponds to the reconstruction loss, optimizing both the encoder and the decoder. The second term refers to the embedding loss, which aims to optimize the embedding space. Lastly, the third term represents the commitment loss, ensuring that the encoder commits to an embedding and its output does not grow.
\vspace{-1mm}
\begin{equation}
\mathcal{L}_{vq}\hspace{-0.3mm}=\hspace{-0.3mm}||s-\hat{s}||^2
+||sg[\mathcal{S}_e(s)]-z_s||_2^2+||sg[z_s]-\mathcal{S}_e(s)||_2^2
\label{eq:loss:vq}
\end{equation}
Here,  $sg[\cdot]$ denotes the stop-gradient operation.

For a given two-dimensional image, we employ a pre-trained Contrastive Language–Image Pretraining (CLIP) model for encoding, subsequently extracting features from the penultimate layer to serve as image attributes. These features are then flattened in a manner akin to language embeddings, adhering to the dimensions of height and width to achieve localized encoding $e_p$.

\subsection{Shape-aware Multimodal Language Model}
\label{sec:method:lm}

\hspace{3mm} To facilitate seamless interaction with large language models, we construct an alignment between shape-aware grammar and pre-trained natural language models, adhering to the learning sequence of ``word-sentence-paragraph'' that mirrors human language acquisition. 

\textbf{From Shape Tokens to Shape Words.}
The input continuous high-dimensional 3D geometries $s$ are transmuted into discrete token sequences $z_s = \{z_s^i\}_{i=0}^{l_s-1}$ via the pre-trained 3D Vector Quantized-Variational AutoEncoder (VQ-VAE). 
Nonetheless, this numerical sequence poses a significant challenge for language models to decipher. 
Consequently, ShapeGPT adopts an explicit approach of substituting token $z_s^i$ (token ID $j$) with the word $w_s^i$ (\textcolor{blue}{<shape\_id\_j>}), for example, shape token ID $8$, $21$, $303$ are replaced with shape words 
\textcolor{blue}{<shape\_id\_8>}, 
\textcolor{blue}{<shape\_id\_21>}, 
\textcolor{blue}{<shape\_id\_303>}.

\textbf{From Shape Words to Shape Sentences.} 
To maintain the high cohesion of shapes, ShapeGPT constructs serialized shape sentences $s_s = \{s_s^i\}_{i=0}^{l_s+1}$ from the discrete shape words $w_s = \{w_s^i\}_{i=0}^{l_s-1}$, delineating them from natural language using the initial symbol 
\textcolor{blue}{<shape\_id\_m>} 
and the terminal symbol 
\textcolor{blue}{<shape\_id\_m+1>}. 
For instance, assuming that $m > 303$, the word sequence consisting of 
\textcolor{blue}{<shape\_id\_8>}, 
\textcolor{blue}{<shape\_id\_21>}, 
\textcolor{blue}{<shape\_id\_303>}
would be transmuted into 
\textcolor{blue}{<shape\_id\_m>}
\textcolor{blue}{<shape\_id\_8>}
\textcolor{blue}{<shape\_id\_21>}
\textcolor{blue}{<shape\_id\_303>}
\textcolor{blue}{<shape\_id\_m+1>}. 
This ensures the integrity of the shape representation while incorporating it into instructions.

\textbf{From Shape Sentences to Shape-aware  Multimodal Language Paragraphs. }
Formally, the shape sentence $s_s$ has aligned the language with the ability to be processed by the language model. We replace the caption placeholder 
\textcolor{orange}{<caption\_place\_holder>} and shape placeholder 
\textcolor{blue}{<shape\_place\_holder>} present in the instruction with the input caption and shape sentence, thereby constructing a coherent natural language paragraph.
Then ShapeGPT utilizes T5~\cite{2020t5} as the language model and tokenizes shape-aware natural language paragraphs into tokens using the WordPiece method with the unigram language model~\cite{kudo2018subword}.
Notably, if an image placeholder 
\textcolor[RGB]{0,128,0}{<image\_place\_holder>} exists in the instruction, ShapeGPT maps the Shape-aware tokens into embeddings, and concatenates them with the image embeddings to form the Shape-aware Multimodal input.

\textbf{Multimodal Language Model.}
ShapeGPT employs T5~\cite{2020t5} as the foundational language model, which is constituted by a transformer-based Language Encoder and a transformer-based Language Decoder, facilitating sequence-to-sequence tasks for uniformly encoded multimodal language paragraphs. Specifically, our multimodal tokens or embeddings are initially mapped to a sequence of embeddings, subsequently, the output sequence is predicted via the Language Encoder and Language Decoder model.

\begin{figure}[t]
	\centering
	\includegraphics[width=\linewidth]{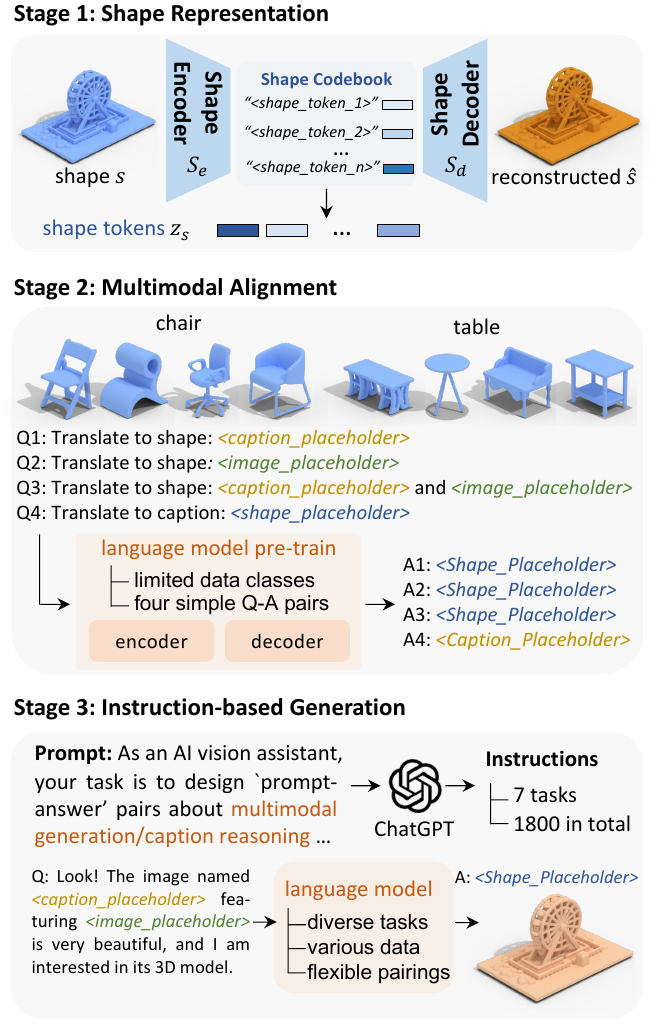}
\vspace{-12pt}
\caption{ Training Scheme. We introduce three training steps for our ShapeGPT (\cref{sec:method: strategy}): First we learn a shape codebook for discrete shape representation. Then we align Vision-Shape-Language models using a mixture of multimodal corpus to comprehend semantic coupling among these modalities. Finally, we fine-tune ShapeGPT with diverse instructions for shape-relevant tasks.}
	\label{fig:training}
 \vspace{-6pt}
\end{figure}

\subsection{Training on Shape-Image-Text Dataset}
\label{sec:method: strategy}

\hspace{3mm} Transitioning natural language models to shape-centric tasks is non-trivial, primarily due to the challenge of enabling the language model to comprehend multimodal language paragraphs.
Therefore, after establishing shape representation, we propose a progressive learning process, initially facilitating the language model's understanding of shape or image vocabulary, followed by proceeding to instruction-based tasks. 
As shown in ~\cref{fig:training}, our training strategy contains three stages: (1) Shape Representation Stage, which involves discretizing continuous shapes and assembling a representative shape corpus. (2) Multimodal Alignment Stage, which aligns the language model with shape-image-text through four fundamental tasks. (3) Instruction-based Generation Stage, which entails constructing instructional generation tasks to achieve more comprehensive functionality.

\textbf{Shape Representation Stage.} ShapeGPT pre-trains a 3D VQ-VAE model, tokenizing shapes into tokens, as demonstrated in Sec.~\ref{sec:method:corpus}. Upon completion of the model's training, akin to the CLIP model responsible for encoding images, it ceases to undergo optimization in the subsequent stages.

\textbf{Multimodal Alignment Stage.}
To facilitate the language model's  initial understanding of Shape-Image-Text vocabulary, as well as the execution of basic instructions, ShapeGPT is  pre-trained on a limited data scope, that is, the chair and table classes with high-quality text annotations~\cite{chen2019text2shape} within the ShapeNet~\cite{shapenet2015} dataset. 
Thus, we devise four simple \textbf{Q-A} pairs to perform Text-to-Shape, Image-to-Shape, Multimodal-to-Shape, and Shape-to-Text tasks, as shown in \cref{fig:training}.

By optimizing the language model via the loss between the generated sequence and the answer sequence, we initially align natural language with the Shape-Image-Text vocabulary. At this stage, although ShapeGPT possesses preliminary shape-centric multimodal generation capabilities, it still exhibits limitations due to monotonous instructions, simplistic tasks, and inadequate data generalization.

\begin{figure*}[t]
	\centering
	\includegraphics[width=\linewidth]{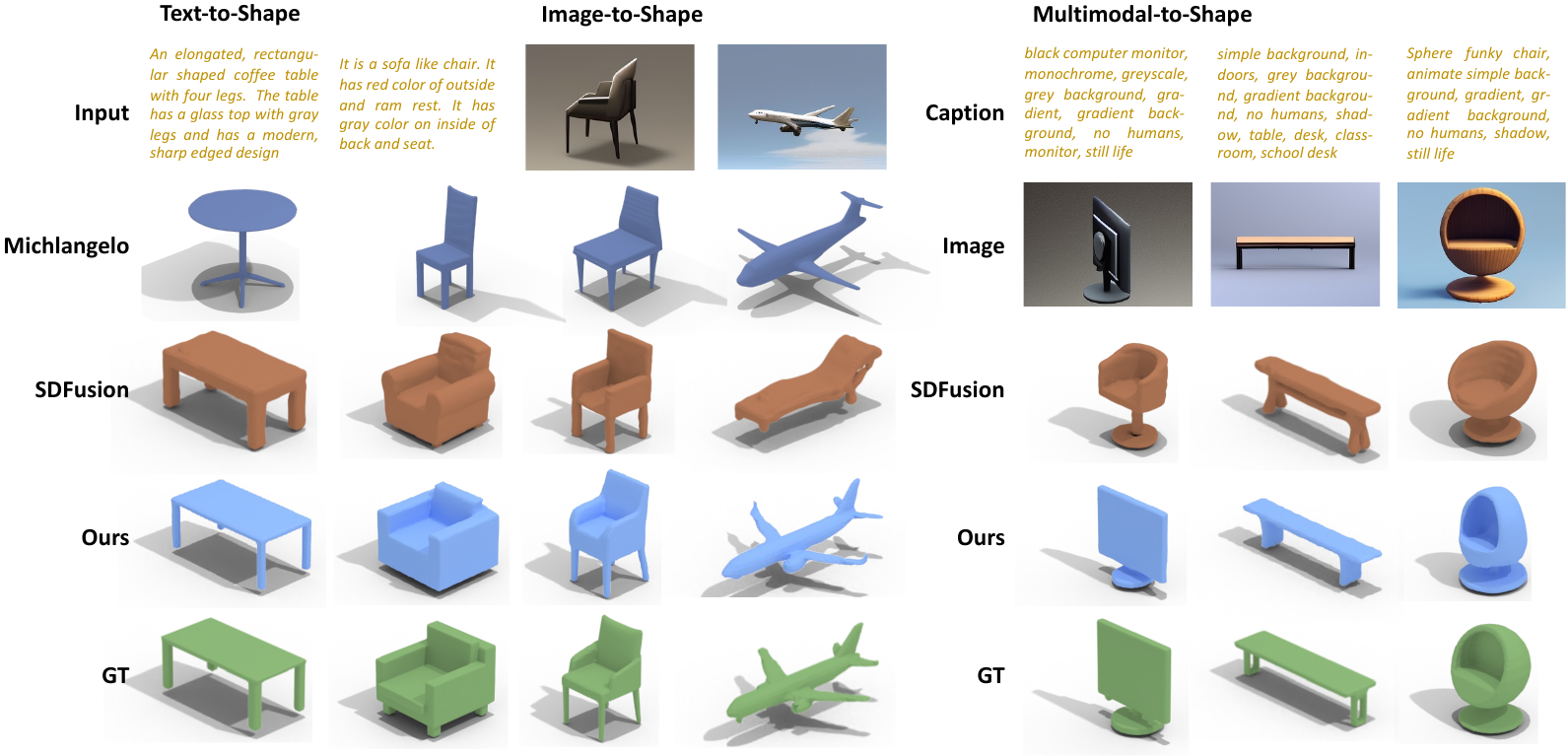}
\vspace{-3pt}
\caption{Qualitative comparison of the state-of-the-art methods. We provide these generated shape results alongside ground truth references from text-to-shape, image-to-shape, and multimodal-to-shape processes (which combine various modalities). We observe that our generated shapes more accurately align with the multimodal prompts. }
	\label{fig:exp}
 \vspace{0pt}
\end{figure*}

\textbf{Instruction-based Generation Stage.}
To achieve comprehensive shape-centric multimodal tasks, we design various prompts and used ChatGPT to construct hundreds of instructions for each task, such as unimodal or multimodal shape generation, shape caption, editing, completion, and inference, etc. 
Additionally, the complexity of the dataset increases significantly, extending to more categories of shapes, as detailed in Sec.~\ref{sec:comp:metric}. 
The constructed commands contain at least one modal placeholder, which can be randomly combined with corresponding modal data~\cref{fig:training}.
The placeholder, including \textcolor{blue}{<shape\_place\_holder>}, \textcolor{orange}{<caption\_place\_holder>} and \textcolor[RGB]{0,128,0}{<image\_place\_holder>}, can be replaced with any set of  caption and image data pair, thus offering remarkable flexibility and substantially enlarging the dataset. 
For instance, in the case of ShapeNet~\cite{shapenet2015}, it can produce data on the order of $10^9$, meeting the training and multi-task generation needs.

\section{Experiments}
\vspace{-3pt}
\label{experiments}

\hspace{3mm} To evaluate the performance of Shape-GPT, comprehensive shape-related multimodal experiments are systematically conducted, implemented, and evaluated on two datasets: ShapeNet and Objaverse (\cref{sec:comp:detail}). 
Quantitative and qualitative evaluations regarding tasks such as text-to-shape generation, shape-to-caption generation, image-to-shape generation, multimodal-to-shape generation, shape editing, shape completion, and shape reasoning have been presented and compared with state-of-the-art methods (\cref{sec:comp:general}). 
Finally, two sets of ablation studies are designed to investigate the influence of the model's structure and parameters (\cref{sec:ablation}).
More experimental results and analyses are available in the supplementary materials.

\subsection{Experimental Setup}
\label{sec:comp:metric}
\textbf{Datasets.} 
We have conducted experiments on ShapeNet~\cite{shapenet2015}, a publicly accessible and challenging shape datasets. 
In accordance with previous methods~\cite{mittal2022autosdf,cheng2022sdfusion,xie2019pix2vox}, we select 16 object categories, encompassing approximately 50,000 models, following Xu's~\cite{xu2019disn} training-test division.
For the task of text-to-shape generation, evaluations are performed on the `chair' and `table' classes, for which the Text2Shape dataset provides detailed text annotations.
As for text-image multi-modalities-to-shape generation, we utilize the Moat~\cite{yang2022moat} method to generate tags, which serve as text annotations for the other classes.

\textbf{Evaluation Metrics.} For shape generation tasks, we mainly use four metrics: 3D Intersection over Union (IoU)~\cite{iou} to measure the overlap between predicted and actual 3D shapes, Chamfer Distance (CD)~\cite{cd} to assess geometric similarity, F-score@1\%~\cite{sokolova2006beyond} to evaluate precision and recall within the top 1\% of results, and ULIP~\cite{xue2022ulip} to assess the perceptual loss between caption and generated shape.
The first three metrics are termed reconstruction metrics, primarily assessing generation with images or partial shapes as one of the conditions. ULIP is named as a generative metrics, focusing on generation with captions as a key condition.
For caption generation tasks, we use pre-trained CLIP~\cite{radford2021clip} model to measure the similarity between the generated text and the text annotation.

\label{sec:comp:detail}
\textbf{Implementation Details.} 
For images, we employed the clip-vit-large-patch14 model to extract features before the final layer, and subsequently designed a six-layer perceiver architecture along with a linear layer to align the image features within the language space. 
For shapes, our codebook size is set at $8192\times64$. For input shapes of $64\times64\times64$, we performed three downsampling operations followed by quantization.
In our language model, we adopt T5 as the foundational architecture. The standard configuration includes a 12-layer transformer for both encoder and decoder. The dimensionality specifications are as follows: feed-forward networks at $d_\text{ff} = 3072$, attention mechanism with an inner dimensionality of $d_\text{kv} = 64$, and other sub-layers and embeddings standardized at $d_\text{model} = 768$.
Moreover, for all stages, we employ the AdamW~\cite{adamw} optimizer and the Cosine Annealing Warm Restarts scheduler strategy with $T_0=5$ and $T_{mult}=2$.
The shape tokenizer model has an initial learning rate of $10^{-4}$, a batch size of 32, and is trained for 315 epochs. 
During the multimodal pre-training stage, the initial learning rate is $4\times10^{-4}$, the batch size is 24, and the model is trained for 635 epochs. 
In the instruction fine-tuning stage, the initial learning rate is $10^{-4}$, the batch size is 24, and the model is trained for 315 epochs.
All models are trained on 4 A100 GPUs.

\begin{figure}[t]
	\centering
	\includegraphics[width=\linewidth]{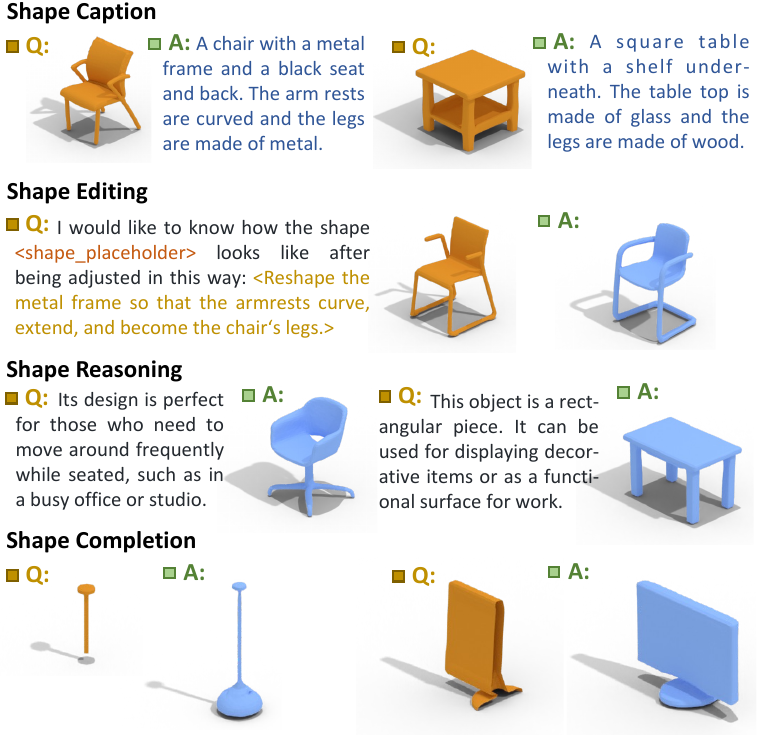}
\vspace{-12pt}
\caption{More results on shape-relevant generation tasks, including shape captioning, editing, reasoning, and completion. The blue shapes and texts are our generation, the orange is the inputs.}
	\label{fig:exp2}
 \vspace{-12pt}
\end{figure}

\subsection{Comparisons on Shape-relevant Tasks}
\label{sec:comp}
\label{sec:comp:general}
\hspace{3mm} By introducing ShapeGPT, a unified multimodal natural language framework capable of comprehending and executing a multitude of shape-related tasks, as shown in ~\cref{fig:exp2}, we then open up the exploration of ShapeGPT through the evaluations of multimodal generation, editing, and inference, among others.
Here we first evaluate ShapeGPT by comparing it with these state-of-the-art works, including Pix2Vox~\cite{xie2019pix2vox}, AutoSDF~\cite{mittal2022autosdf}, SDFusion~\cite{cheng2022sdfusion}, and Michelangelo~\cite{zhao2023michelangelo} on common shape-relevant tasks, including text-to-shape, shape-to-text, image-to-shape, and multi-modal-to-shape. 
Subsequently, for directive-based tasks such as editing, inference, and completion, we will showcase qualitative outcomes.
Experimental findings indicate that ShapeGPT, through a unified framework, is capable of accomplishing numerous shape-related tasks. Concurrently, in terms of metrics for common tasks, it exhibits comparability with dedicated models. Additional experimental results can be found in the supplementary materials.

\label{sec:comp:text}
\textbf{Comparisons on Image-to-Shape.} 
For the task of image-based shape generation, the aim is to reconstruct the corresponding shape from a single rendered image. 
Evaluations are conducted on the ShapeNet~\cite{shapenet2015} dataset, following the training-test partitioning proposed by Xu et al.~\cite{xu2019disn}, and comparisons are made with the latest methods like autoregressive prior-driven AutoSDF, diffusion-centric SDFusion, and the latent alignment-oriented Michelangelo. 
Reconstruction metrics are used for evaluation, similar to previous methods, providing results for the chair class and the average across all classes.
~\cref{tab:tm:comp:i2s} shows that ShapeGPT is comparable to the latest methods, and qualitative results ($cf.$~\cref{fig:exp}) also demonstrate its ability to generate realistic shapes.

\textbf{Comparisons on Text-to-Shape.} 
Generating shapes from textual descriptions involves creating semantically consistent shapes based on provided shape captions. 
The efficacy of ShapeGPT on this task is evaluated on the ShapeNet dataset. 
Owing to the limitation of ShapeNet's~\cite{shapenet2015} caption annotations, only covering chairs and tables as provided by Text2shape~\cite{chen2019text2shape}, and considering that ShapeGPT embodies a multi-task unified framework, tags are employed as auxiliary elements for caption-lacking classes to ensure uniform form. 
To ensure a fair comparison, we report quantitative generation results for chairs and tables in ~\cref{tab:tm:comp:t2s}, while also showcasing qualitative results in ~\cref{fig:exp}. 
The results indicate that our method is close to the previous optimal approach, which is attributable to the powerful linguistic comprehension capacity of the language model, facilitating effective perception of shape-relevant features from captions.

\begin{table}[t]
\centering
\resizebox{0.9\columnwidth}{!}{%
\begin{tabular}{ccccccc}

\toprule
\multirow{2}{*}{Methods} &
\multicolumn{3}{c}{the chair category}&
\multicolumn{3}{c}{all categories}\\
\cmidrule(lr){2-4} \cmidrule(lr){5-7} 

& IoU$\uparrow$ & CD$\downarrow$ & F-score$\uparrow$ & IoU$\uparrow$ & CD$\downarrow$ & F-score$\uparrow$\\
\toprule
Pix2Vox~\cite{xie2019pix2vox} &0.467 &2.521& 0.335 &-&-&-\\
AutoSDF~\cite{mittal2022autosdf}& 0.577 &1.331 &0.414 &-&-&-\\
SDFusion~\cite{cheng2022sdfusion}& \textbf{0.595} & 1.323&0.412&0.532&1.375&0.353\\
Michelangelo~\cite{zhao2023michelangelo}& 0.566 & 1.428& 0.392&0.561&1.374&\textbf{0.402}\\
ShapeGPT (Ours)&  0.593 & \textbf{1.221}& \textbf{0.424}&\textbf{0.570}&\textbf{1.297}&0.396\\
   \bottomrule
\end{tabular}
}
\vspace{-2mm}
\caption{Quantitative Assessment for Image-to-Shape. We evaluate the performance of generating shapes from a single image and report the reconstruction metrics on the chair class and all classes from  ShapeNet dataset. ShapeGPT exhibits superior performance across multiple indices, demonstrating its effectiveness in this task.}
\label{tab:tm:comp:i2s}
\end{table}

\begin{table}[t]
\centering
\resizebox{0.40\columnwidth}{!}{%
\begin{tabular}{cc}
\toprule
Methods &  ULIP$\uparrow$\\
\cmidrule(lr){1-1} \cmidrule(lr){2-2}
SDFusion~\cite{cheng2022sdfusion}& 0.105\\
Michelangelo~\cite{zhao2023michelangelo}& \textbf{0.165}\\
ShapeGPT (Ours)& 0.149\\
   \bottomrule
\end{tabular}%
}
\vspace{-2mm}
\caption{Quantitative Assessment for Text-to-Shape. We evaluate the performance of generating shapes from captions and report generation metrics for the high-quality annotated categories of chairs and tables. ShapeGPT is close to the optimal method and can handle the task of text to shape generation.}
\label{tab:tm:comp:t2s}
\end{table}

\begin{table}[t]
\centering
\resizebox{0.78\columnwidth}{!}{%
\begin{tabular}{ccccc}
\toprule
Methods & IoU$\uparrow$ & CD$\downarrow$ & F-score$\uparrow$ & ULIP$\uparrow$\\

\cmidrule(lr){1-1} \cmidrule(lr){2-5}
SDFusion~\cite{cheng2022sdfusion}& 0.562 & 1.332&0.380&0.172\\
ShapeGPT (Ours)&  \textbf{0.587} & \textbf{1.256}& \textbf{0.402}&\textbf{0.189}\\
   \bottomrule
\end{tabular}%
}
\vspace{-2mm}
\caption{Quantitative Assessment for Multi-modal-to-Shape. The efficacy of generating shapes from combined captions and single images is assessed, and the corresponding generation and reconstruction metrics is reported.
The performance of ShapeGPT excels across all metrics compared to the latest methods in the multi-modal shape generation task.
}
\label{tab:tm:comp:m2s}
\end{table}

\textbf{Comparisons on Multi-modal-to-Shape.} 
We evaluate the capability of generating shapes from multimodal data, using a single rendered image and caption. 
As a unified framework, the setup for both image and caption aligns with the Image-to-Shape and Text-to-Shape tasks. 
We conduct a comparative analysis with the state-of-the-art SDFusion~\cite{cheng2022sdfusion} method, which supports multimodal input, as shown in ~\cref{tab:tm:comp:m2s} and ~\cref{fig:exp}. 
The results indicate that our ShapeGPT outperforms previous methods. We transform different modalities into linguistic sequences, subsequently processed through a language model to formulate sequences emblematic of shapes. This unified representational and processing approach potentially augments the perceptual and integrative efficacy for shape generation.

\begin{table}[t]
\centering
\resizebox{1\columnwidth}{!}{%
\begin{tabular}{cccccccccc}
\toprule

\multicolumn{1}{c}{Shape caption }&
\multicolumn{3}{c}{Shape completion }&
\multicolumn{3}{c}{Shape reasoning}&
\multicolumn{3}{c}{Shape editing}\\
\cmidrule(lr){1-1} \cmidrule(lr){2-4} \cmidrule(lr){5-7} \cmidrule(lr){8-10}
CLIP$\uparrow$
& IoU$\uparrow$ & CD$\downarrow$ & F-score$\uparrow$ 
& IoU$\uparrow$ & CD$\downarrow$ & F-score$\uparrow$ 
& IoU$\uparrow$ & CD$\downarrow$ & F-score$\uparrow$ 

\\ 
\toprule

0.812& 
0.622& 1.155& 0.477&
0.292& 2.328& 0.272&
0.534& 1.405& 0.367\\
\bottomrule
\end{tabular}%
}
\vspace{-1mm}
\caption{Quantitative results of ShapeGPT on more shape-related tasks, including shape captioning, shape completion, shape reasoning, and Shape editing.}
\vspace{-2mm}
\label{tab:tm:comp:m2t}
\end{table}

\label{sec:comp:m2t}
\textbf{Evaluations on Additional Tasks.}
ShapeGPT is not limited to shape generation tasks, but can also execute a variety of shape-centric tasks based on instructions, including shape caption, shape completion, shape reasoning, and shape editing, as shown in \cref{fig:exp2} and \cref{tab:tm:comp:m2t}.

\textit{Shape caption} aims to generate a descriptive sentence based on the input shape. We evaluate the performance on the chair and table classes, and calculate the similarity between the generated caption and the high-quality annotation with the pre-trained CLIP~\cite{radford2021learning} model. 
\textit{Shape completion} involves reconstructing a complete shape from the given partial one. We randomly remove 0\% to 50\% from original models in the ShapeNet dataset, followed by ShapeGPT's reconstruction from the remaining parts, and comparing with intact models using reconstruction metrics. 
\textit{Shape reasoning}, distinct from the text-to-shape task, generates shapes based on descriptions of function, application, or user needs, strictly avoiding direct inputs regarding the classification or explicit appearance of the shape. We evaluate this task by calculating the reconstruction metrics between the generated shapes and the target shapes. 
In \textit{Shape editing}, based on the original shape and editing instructions, ShapeGPT creates a new shape that should minimally change the original while following the instructions.
We calculate the reconstruction metrics on the chairs and tables classes, where the editing instructions from the original to the target shape are generated by ChatGPT.

The experimental results indicate that, benefiting from a unified sequence-to-sequence language encoding format, our ShapeGPT seamlessly integrates and executes multiple multimodal shape-related tasks within a single architecture, devoid of any discernible transitional processes, and demonstrates commendable performance across these varied tasks.

\begin{table}[h]
\centering
\resizebox{0.62\columnwidth}{!}{%
\begin{tabular}{cccc}
\toprule
Methods & IoU$\uparrow$ & CD$\downarrow$ & F-score$\uparrow$\\

\cmidrule(lr){1-1} \cmidrule(lr){2-4}

64 tokens &0.473 &2.043& 0.315 \\
T5-small& 0.558 &1.362 &0.389 \\
Without pre-training& 0.547 &1.333 &0.374 \\
\cmidrule(lr){1-1} \cmidrule(lr){2-4}
ShapeGPT (Ours)&\textbf{0.570}&\textbf{1.297}&\textbf{0.396}\\
   \bottomrule
\end{tabular}%
}
\vspace{-1mm}
\caption{Ablation Study. We explore three aspects: varying sequence lengths, model sizes, and the necessity of pretraining. The results indicate that 512 tokens, the T5-base model, and the shape-aware pretraining approach are most effective given the current computational resources.}
\label{tab:abs}
\end{table}

\subsection{Ablation Studies}
\label{sec:ablation}

\hspace{3mm} ShapeGPT initially discretizes input Shapes into \textit{fixed-length} tokens. Then shape-aware multimodal language paragraphs are processed by the \textit{T5-base model}~\cite{raffel2020t5}. 
The T5 model is \textit{first pre-trained} on partial data and simple instructions, before expanding to full dataset and complex instructions. 
Our ablation studies focuses on the shape token length, the language model size, and the necessity of pre-training in the image-to-shape task, as shown in ~\cref{tab:abs}.

\textbf{Assessment of Shape Token Length.}  In this study, the shape token length is set to 512 (flattened from an 8×8×8 grid), and we also test the length of 64 (flattened from a 4×4×4 grid), with larger voxel configurations constrained by GPU memory.  Experiments show that 512 tokens offers better performance compared to shorter shape sequences.

\label{sec:ablation:motionsize}
\textbf{Model size.} We evaluate the T5-small and T5-base models~\cite{raffel2020t5}, which have 60M and 220M parameters, respectively. Larger models are not feasible due to the GPU memory limitation. Experiments demonstrate that T5-base achieves better results than T5-small, while previous comparisons illustrate that the T5-base model basically meets the requirements of various shape-related tasks.

\label{sec:ablation:strategy}
\textbf{The Necessity of Pre-training. }
Quantitative results demonstrate that pre-training significantly enhances model performance compared to models without it.
We observe that outputs from models without pre-training often fail to conform to the expected modality or coalesce into coherent shape language, illustrating the importance of pre-training in enhancing the model's multimodal language alignment and the learning of shape language grammar and syntax.

\section{Conclusion and Limitation}

\textbf{Limitation.}
In anticipation of overcoming the constraints of this study, we plan to endow ShapeGPT with more capabilities, including more shape-centered tasks, generating textured shapes, and supporting more modalities such as voice and video. 
Moreover, the theoretical versatility of our language construction methodology and training strategy holds promise for broader applications, and we look forward to applying it in more fields, such as multi-object, 3D scenes, dynamic shape, and video generation.

\textbf{Conclusion.}
In this paper, we introduce ShapeGPT, a shape-centric multi-modal large language model, establishes a unified architecture for various shape-related tasks based on instructions. The core of our approach is to intermix instructions with multi-modal inputs such as shapes, captions, and images  into a natural language format, thereby facilitating sequence-to-sequence generation via the large language model. 
ShapeGPT demonstrates the ability to perform various shape-related tasks,  while quantitative evaluations reveal that its performance in common shape generative tasks is comparable to the state-of-the-art methods.

\newpage

{
    \small
    \bibliographystyle{ieeenat_fullname}
    \bibliography{main}
}

\end{document}